\documentclass[letterpaper]{article} 
\usepackage{aaai25}  
\usepackage{times}  
\usepackage{helvet}  
\usepackage{courier}  
\usepackage[hyphens]{url}  
\usepackage{graphicx} 
\usepackage{natbib}  
\usepackage{caption} 
\usepackage{algorithm}
\usepackage{algorithmic}
\usepackage{bm}
\usepackage{booktabs}
\usepackage{amsmath}
\usepackage{amssymb}
\usepackage{xcolor}

\usepackage{newfloat}
\usepackage{listings}
\DeclareCaptionStyle{ruled}{labelfont=normalfont,labelsep=colon,strut=off} 
\floatstyle{ruled}
\newfloat{listing}{tb}{lst}{}
\floatname{listing}{Listing}
\usepackage{standalone}
\usepackage{cite}

\title{TrackGo: A Flexible and Efficient Method for Controllable Video Generation}
\author{
    Haitao Zhou\equalcontrib\textsuperscript{\rm 1, \rm 2}, \
    Chuang Wang\equalcontrib\textsuperscript{\rm 1, \rm 2}, \
    Rui Nie \textsuperscript{\rm 1}, \
    Jinlin Liu \textsuperscript{\rm 2}, \
    Dongdong Yu \textsuperscript{\rm 2}, \\
    Qian Yu \textsuperscript{1 \textdagger}, \
    Changhu Wang \textsuperscript{2 \textdagger} 
}

\affiliations{
    \textsuperscript{\rm 1} Beihang University, \textsuperscript{\rm 2} AIsphere Tech\\
    \{zhouhaitao, chuangwang, qianyu, nierui\_\}@buaa.edu.cn \\
    \{yudongdong, linjinxiao, wangchanghu\}@aishi.ai \\
    https://zhtjtcz.github.io/TrackGo-Page/
}

\begin{document}

\maketitle

\begin{abstract}

Recent years have seen substantial progress in diffusion-based controllable video generation. 
However, achieving precise control in complex scenarios, including fine-grained object parts, sophisticated motion trajectories, and coherent background movement, remains a challenge. 
In this paper, we introduce \textit{TrackGo}, a novel approach that leverages free-form masks and arrows for conditional video generation. 
This method offers users with a flexible and precise mechanism for manipulating video content. 
We also propose the \textit{TrackAdapter} for control implementation, an efficient and lightweight adapter designed to be seamlessly integrated into the temporal self-attention layers of a pretrained video generation model. This design leverages our observation that the attention map of these layers can accurately activate regions corresponding to motion in videos.
Our experimental results demonstrate that our new approach, enhanced by the TrackAdapter, achieves state-of-the-art performance on key metrics such as FVD, FID, and ObjMC scores.

\end{abstract}

\section{Introduction}

With the rapid development of diffusion models~\cite{ddpm_ho_2020, song2020denoising, dhariwal2021diffusion, song2020score}, video generation has witnessed significant progress, with the quality of generated videos continuously improving. Unlike text-based~\cite{blattmann2023align, zhang2023i2vgen, guo2023animatediff} or image-based~\cite{blattmann2023stable} video generation, controllable video generation~\cite{hu2023animate, yin2023dragnuwa, wu2024draganything} focuses on achieving precise control over object movement and scene transformations in generated videos. This capability is particularly valuable in industry such as film production and cartoon creation.

\begin{figure}[t!]
\centering
\includegraphics[width=0.46\textwidth]{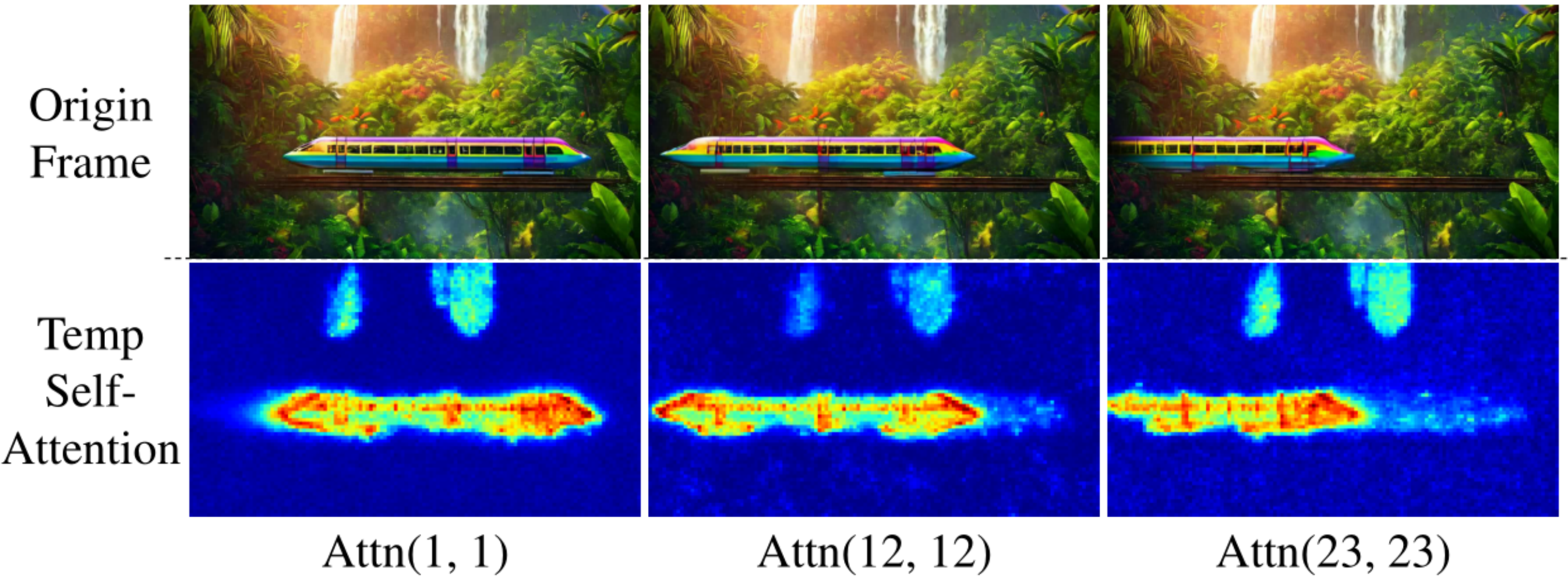}
 \caption{
    \footnotesize Attention map visualization of the last temporal self-attention layer in Stable Video
Diffusion Model. The highlighted areas in the attention map correspond to the moving areas in the video. The video has a total of 25 frames, and we selected frames 1, 12, and 23 at equal intervals for visualization. And $Attn(i,j)$ denotes the temporal attention map between frame $i$ and frame $j$.
}
\label{fig:hatmap}
\end{figure}

\begin{figure*}[htb]
\centering
\includegraphics[width=0.98\textwidth]{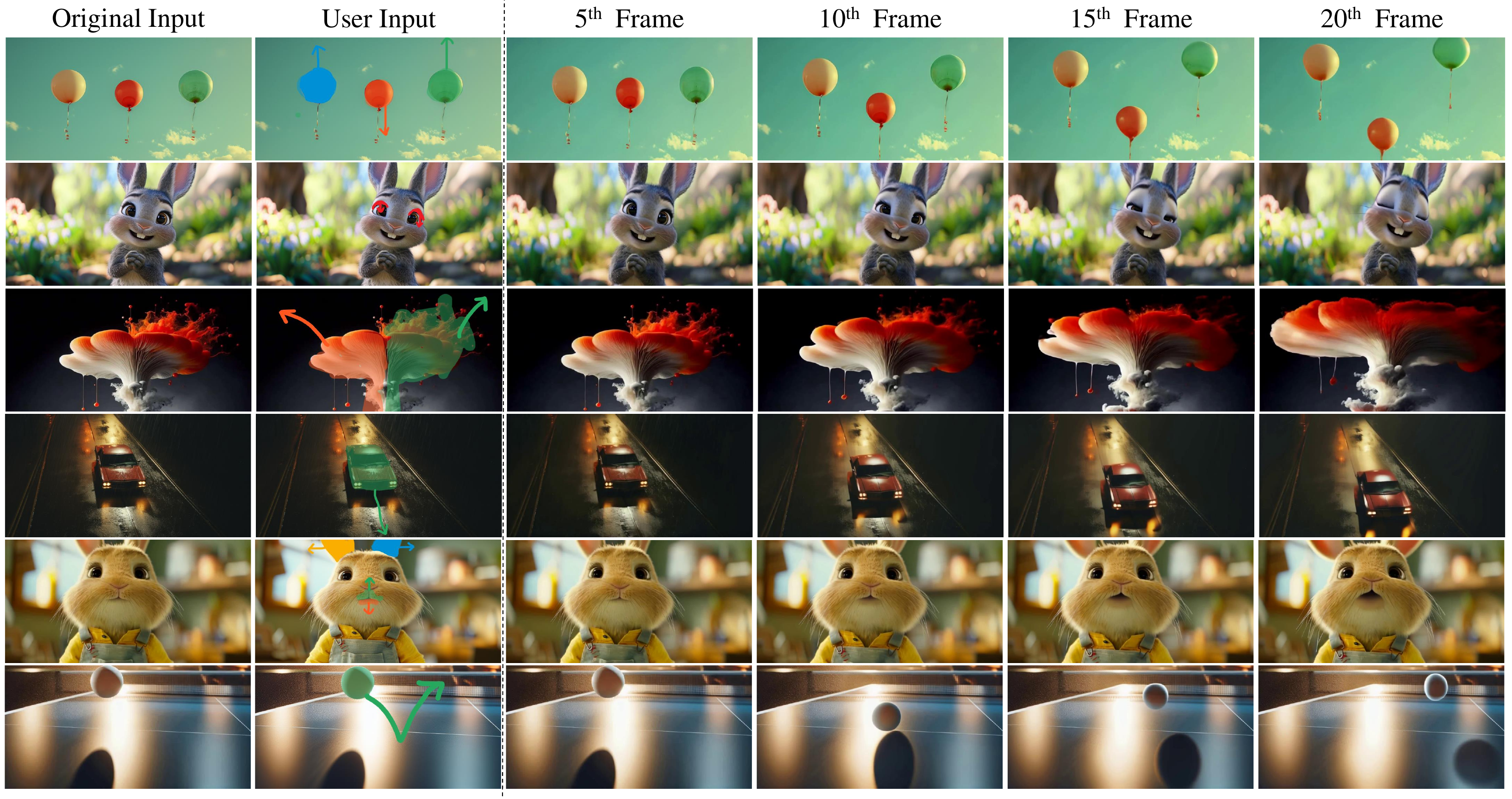}
 \caption{
    Example videos generated by our proposed \textit{TrackGo}. Given an initial frame, users specify the target moving object(s) or part(s) using free-form masks and indicate the desired movement trajectory with arrows. TrackGo is capable of generating subsequent video frames with precise control. It can handle complex scenarios that involve multiple objects, fine-grained object parts, and sophisticated movement trajectories.
    }
  \label{fig:tears}
\end{figure*}

Controllable video generation remains a highly challenging task. The primary challenge is precise control, which includes managing the target movement objects and their trajectories. Existing methods often struggle to achieve precise control over these elements. For instance, DragAnything \cite{wu2024draganything} employs a center point and a Gaussian map to guide the target object along a predefined path. However, it fails to control the movement of partial or fine-grained objects effectively.
Another approach, Boximator \cite{wang2024boximator}, utilizes bounding boxes to dictate motion control. It uses a box to specify the target area, where the sequence of movements of the box guides the motion of the target. Unfortunately, bounding boxes often encompass redundant regions, which can interfere with the motion of the target and disrupt the coherence of the background in the generated videos.
The second challenge is efficiency. Existing works often incorporate conditions in a way that significantly increases the number of model's parameters.
For instance, DragAnything utilizes the architecture of ControlNet~\cite{zhang2023adding}, and DragNUWA~\cite{yin2023dragnuwa} employs heavy encoders to map guidance signals into the latent space of the pretrained model. These design choices inevitably lead to slower inference times, which can impede the practical deployment of these models in real-world applications.

In this work, we tackle the task of controllable video generation by addressing two crucial questions: First, \textit{what} type of control should be employed to accurately describe the motion of the target? Second, \textit{how} can this control be implemented efficiently?

For the \textit{first} question regarding the type of control suitable for describing the motion of the target, we propose a novel combination of a free-form mask and an arrow to guide motion. Specifically, users can define the target area with a brush, allowing for precise specification ranging from entire objects to partial areas. The trajectory of the movement is indicated by an arrow, also drawn by the user, which provides clear directional guidance.
For the \textit{second} question concerning efficient control implementation, we introduce a novel approach that involves injecting conditions into the temporal self-attention layers. We observed that the attention maps generated by these layers effectively highlight areas of motion within a video, a finding also supported by previous research \cite{ma2023trailblazer}. As demonstrated in Fig.\ref{fig:hatmap}, the region of a moving train is distinctly activated in the attention map of the temporal self-attention layer.
Building on this insight, we propose directly manipulating the attention map of the temporal self-attention layer to achieve precise motion control. This method not only enhances accuracy but also minimizes additional computational overhead.

We introduce \textbf{TrackGo}, a novel framework for controllable video generation that leverages user inputs to direct the generation of video sequences. TrackGo uses free-form masks and arrows provided by users to define target regions and movement trajectories, respectively, as illustrated in Fig.~\ref{fig:tears}. This approach consists of two stages: Point Trajectories Generation and Conditional Video Generation.
In the first stage, TrackGo automatically extracts point trajectories from the user-defined masks and arrows. These trajectories serve as precise blueprints for video generation.
In the second stage, we use the Stable Video Diffusion Model (SVD)~\cite{blattmann2023stable} as the base model, accompanied by an encoder that encodes the motion information.
To ensure that the guidance is precisely interpreted by our model, we introduce the novel \textbf{TrackAdapter}. This adapter effectively modifies the existing temporal self-attention layers of a pre-trained video generation model to accommodate new conditions, enhancing the model's control over the generated video.

Specifically, the TrackAdapter introduces a dual-branch architecture within the existing temporal self-attention layers. It integrates an additional self-attention branch running parallel to the original. 
This new branch is specifically designed to focus on the motion within the target area, ensuring that the movement dynamics are captured with high fidelity. Meanwhile, the original branch continues to handle the rest areas. 
This architecture not only ensures accurate and cohesive generation of both the specific movements of the target and the overall video context but also modestly increases the computational cost.
Furthermore, we introduce an attention loss to accelerate model convergence, thereby enhancing efficiency.
This balance between control fidelity and efficiency is crucial for practical applications of video generation.

In summary, our contributions are three fold:
\begin{itemize}
\item We introduce a novel motion-controllable video generation approach named TrackGo. 
This method offers users a flexible mechanism for motion control, combining masks and arrows to achieve precise manipulation in complex scenarios, including those involving multiple objects, fine-grained object parts, and sophisticated movement trajectories.
\item A new component, the TrackAdapter, is developed to integrate motion control information into temporal self-attention layers effectively and efficiently.
\item We conduct extensive experiments to validate our approach. The experimental results demonstrate that our model surpasses existing models in terms of video quality (FVD), image quality (FID), and motion faithfulness (ObjMC). 
\end{itemize}

\section{Related Work}

\begin{figure*}[t]
  \centering
  \includegraphics[width=0.98\textwidth]{ 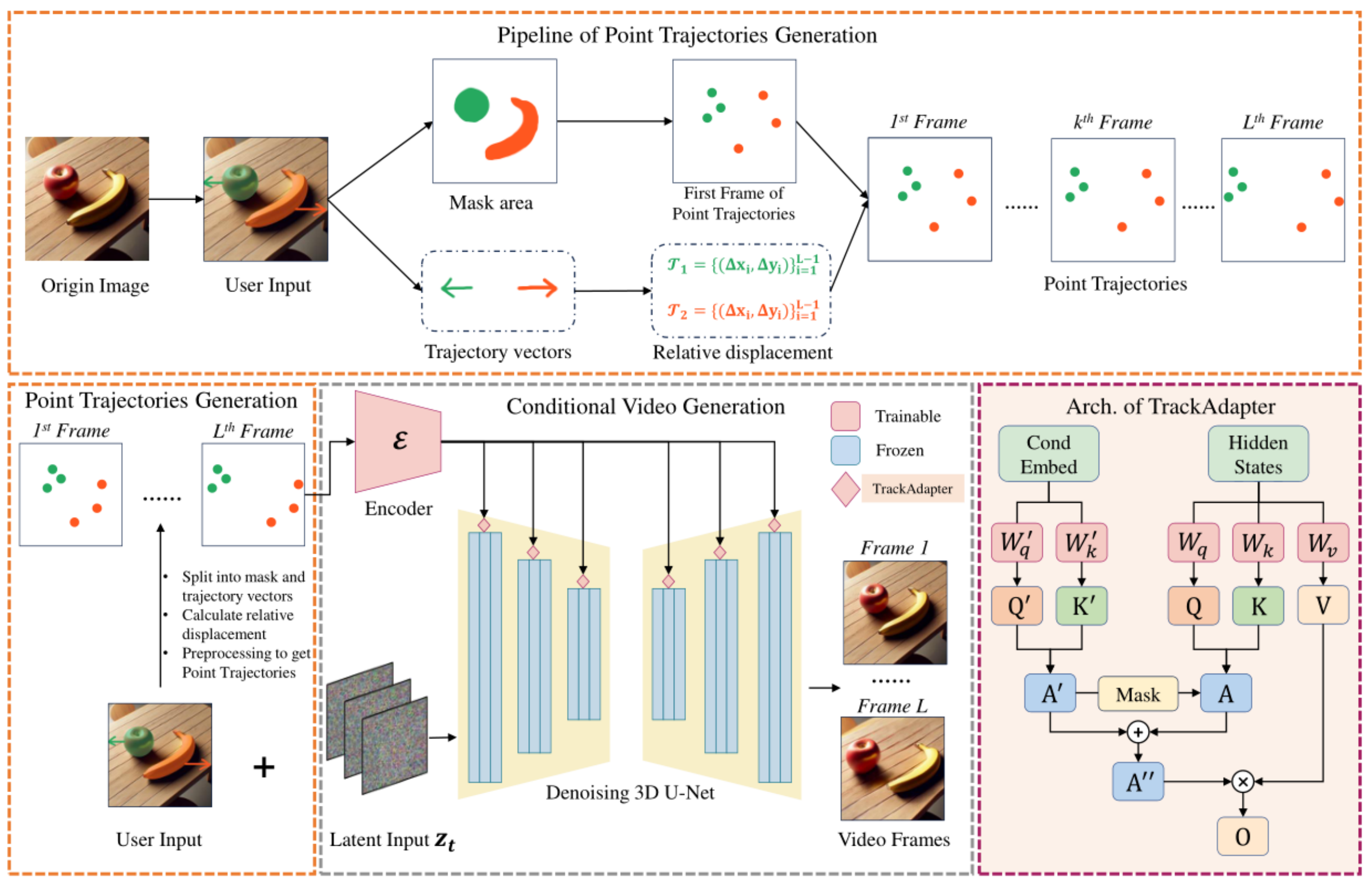}
  \caption{\textbf{Top: Pipeline of Point Trajectories Generation.} User's inputs are divided into masks and trajectory vectors for processing. Each mask corresponds to a trajectory vector. For each mask area, $K*s$ points are randomly selected. The trajectory vector is then subdivided by the frame number to attain the relative displacement $\mathcal{T}$ of each point between frames between adjacent frames. The final step is to combine this relevant data to construct point trajectories.
  \textbf{Bottom: Overview of TrackGo}. TrackGo generates videos by taking user input $\bm{I}$ and latent input $\bm{z}_t$ as inputs based on an image-to-video diffusion model.  Through the pipeline of point trajectories generation, point trajectories $\bm{P}$ can be obtained from $\bm{I}$.  Then the point trajectories $\bm{P}$ are passed through the Encoder $\mathcal{E}$ and injected into the model via the TrackAdapter. $\textbf{Architecture of TrackAdapter}$ describes the calculation process of TrackAdapter.}
  \label{fig:arch}
\end{figure*}

\subsection{Diffusion Model-based Image and Video Generation}
Diffusion models \cite{ddpm_ho_2020} have made great progress in the field of text-to-image generation \cite{betker2023improving,  peebles2023scalable, rombach2022high, saharia2022photorealistic, zhang2023adding, xing2023dynamicrafter}, which directly promotes the progress of basic video diffusion, and many excellent works \cite{chen2024pixart, henschel2024streamingt2v, ho2022video,  zhang2023i2vgen, wang2023modelscope, zeng2023make} have emerged. Early works such as VLDM~\cite{blattmann2023align} and AnimateDiff \cite{guo2023animatediff} 
try to insert temporal layers to complete the generation of text to video. Recent models benefit from the stability of diffusion-based trained models. I2VGen-XL \cite{zhang2023i2vgen}, Stable Video Diffusion (SVD) \cite{blattmann2023stable} achieve surprising results on text-to-video generation and image-to-video, respectively, with large-scale high-quality data. 
While these models are capable of producing high-quality videos, they primarily depend on coarse-grained semantic guidance from text or image prompts, which can lead to actions that do not align with the user's intentions.

\subsection{Controllable Image and Video Generation}

In pursuit of enhancing controllability in image and video content generation, numerous recent studies \cite{mou2024t2i, ye2023ip,  khachatryan2023text2video, ceylan2023pix2video} have integrated diverse methodologies to incorporate additional forms of guidance. Notably, Disco~\cite{wang2023disco}, MagicAnimate~\cite{xu2023magicanimate}, DreamPose~\cite{karras2023dreampose}, and Animate Anyone~\cite{hu2023animate} have each adopted pose-directed approaches, enabling the creation of videos featuring precisely prescribed poses. These advancements reflect a concerted effort towards achieving finer-grained manipulation and tailored video synthesis through pose-guided techniques. Despite demonstrating remarkable performance, these techniques are specifically tailored to and thus limited in application to videos depicting human subjects.

To broaden the scope of controllable video synthesis ~\cite{wang2024videocomposer, guo2023sparsectrl, ma2024follow, wang2023motionctrl, yin2023dragnuwa, wang2024boximator, wu2024draganything, tu2023motioneditor, jiang2023videobooth, qi2023fatezero} and enhance its general applicability, several approaches incorporate control signals into pre-trained video diffusion models. The work of AnimateDiff endeavors,  to integrate LoRA (Low-Rank Adaptation)~\cite{hu2021lora} within temporal attention mechanisms to grasp camera motion learning. Nevertheless, LoRA's potential shortcomings emerge when tasked with thoroughly manipulating and synthesizing sophisticated movements, as its learning capacity and predictive prowess might be constrained by repetitive motion patterns present in the training dataset. Innovations like MotionCtrl~\cite{wang2023motionctrl} and DragNUWA~\cite{yin2023dragnuwa} encode sparse optical flow into dense optical flow as guidance information to inject into the diffusion model to control object motion. Boximator~\cite{wang2024boximator}, on the other hand, pioneers motion regulation by correlating objects with bounding boxes, capitalizing on the model's innate tracking capabilities. Fundamentally, this approach employs dual trajectory sets defined by the bounding box's upper-left and lower-right vertices. A caveat arises from the boxes occasionally misinterpreting background details as part of the foreground, which can inadvertently taint output accuracy. DragAnything~\cite{wu2024draganything} has introduced an approach that employs masks to pinpoint a central point, subsequently generating a Gaussian map for tracking this center to produce a guiding trajectory for model synthesis. It's crucial to note, however, that not all masked areas denote regions of motion. And relying on the ControlNet~\cite{zhang2023adding} structure makes it difficult to achieve satisfactory results in terms of efficiency.

\section{Methodology}
\subsection{Overview}

Our task is motion-controllable video generation. For an input image $\bm{I} \in \mathbb{R}^{H \times W \times 3}$,
and point trajectories $ \bm{P} \in \mathbb{R}^{L \times H \times W \times 3}$ extracted from arrows describes the  trajectories information, generating a video $ \bm{V} \in \mathbb{R}^{L \times H \times W \times 3}$ in line with the trajectories, where $L$ is the length of video.

We use Stable Video Diffusion Model (SVD)~\cite{blattmann2023stable} as our base architecture. The SVD model, similar to most video latent diffusion models~\cite{blattmann2023align}, adds a series of temporal layers on the U-Net of image diffusion to form 3D U-Net. On this basis, We pass the point trajectories $ \bm{P}$  through a trainable encoder $\mathcal{E}$ to obtain a compressed representation $f$. This representation is then injected into each temporal self-attention module, using down-sampling to process and adapt it to the appropriate resolution. We introduce the \textbf{TrackAdapter} and at each temporal self-attention of SVD, a TrackAdapter is added to inject $f$, as shown in Fig.~\ref{fig:arch}.

In the following sections, we will cover three main topics: (1) The advantages of point trajectories and how we obtain and use them. (2) The structure of TrackAdapter and how it helps SVD to understand complex motion patterns and complete the generation of complex actions. (3) The process of training and inferring our model.

\subsection{Point Trajectories Generation}
\label{sec:prepare}

In inference, when the user provides the first frame picture, the masks of editing area, and corresponding arrows. Our approach can convert the masks and arrows into point trajectories $\bm{P}$ through preprocessing, as shown in Fig.~\ref{fig:arch}.

For the training process, we first use DEVA~\cite{cheng2023tracking} to segment the main components in the ground truth video $\bm{P}$ and obtain the corresponding segmentation sequence $\bm{M}_i^j$, where $i$ denotes the $i$-th frame and $j$ denotes the $j$-th component. Then, for the mask sequence $ \{\bm{M_{1}^{j}}\}_{j=1}^{s} $ of the first frame, we need to select $K$ points in each mask area as control points, where $s$ denotes the number of components.
For mask $\bm{M}_i^j$, we randomly select $3K$ points in the white area, and then use the K-means~\cite{macqueen1967some} to obtain $K$ points. This guarantees the uniformity of the chosen $k$ points without incurring too much time overhead. We will have a total of $s * K$ control points after this stage. After obtaining the control points, we use the Co-Tracker~\cite{karaev2023cotracker} to track these points and obtain the corresponding motion trajectories $\mathcal{T} = \{ \{ (x_{i}^{1}, y_{i}^{1}) \}_{i=1}^{L}, \{ (x_{i}^{2}, y_{i}^{2}) \}_{i=1}^{L}, ..., \{ (x_{i}^{s*K}, y_{i}^{s*K}) \}_{i=1}^{L}  \}$, where $L$ is the length of video. Finally, we assign a color to the control points corresponding to the same component, plot the trajectory $\mathcal{T}$ and get point trajectories $\bm{P} \in \mathbb{R}^{L \times H \times W \times 3} $. We built the training dataset using this method, and after data cleaning and other operations, we finally got 110k ($\bm{V}$, $\bm{P}$, $ \{\bm{M_}{1}^{j}\}_{j=1}^{s}$) triple pairs. 

\subsection{Injecting Motion Conditions via TrackAdapter
}
\label{sec:adapter}

\subsubsection{Motion Conditions Extraction.}
We use the same encoder structure from Animate  Anyone~\cite{hu2023animate} to extract the timed features. This process can be obtained from Eq.~\ref{eq:encoder}, where $f$ is the compressed temporal representation of $\bm{P}$. 

\begin{equation}
f = \mathcal{E}(\bm{P})
\label{eq:encoder}
\end{equation}

The $f$ will be down-sampled according to the resolution of different temporal self-attention layers and aligned with its input size.

\subsubsection{TrackAdapter Design.}

In order to use the compressed temporal representation $f$ to guide the model to generate a video corresponding to this action, a straightforward approach is to construct the attention map shown in Fig.~\ref{fig:hatmap} using $f$.
Therefore, we propose a lightweight and simple structure called TrackAdapter. The function of TrackAdapter is to activate a motion region corresponding to a specified object, thus guiding the model generation process.
When injecting point trajectories, TrackAdapter is responsible for activating the motion region of the specified object. We first compute the attention map $\bm{A^{'}}$ for the TrackAdapter:
\begin{equation}
\bm{A}^{'} = \mathrm{softmax}(
\frac{\bm{Q}^{'}(\bm{K}^{'})^{T}}{\sqrt{d}}
)
\label{con:score_p}
\end{equation}
\noindent where $\bm{Q}^{'} = f\bm{W}_{q}^{'}, \bm{K}^{'} = f\bm{W}_{k}^{'}$ are the query, key matrices of the TrackAdapter, $f$ is the compressed representation of $\bm{P}$, obtained from Eq.~\ref{eq:encoder}.

To avoid the impact of the origin temporal self-attention branch on the final active region, we obtain an attention mask according to the attention map to suppress the areas activated by the origin temporal self-attention branch.

We transform the attention map $\bm{A}^{'}$ into the corresponding attention mask $\bm{A}_M$ by setting a threshold $\alpha$:
\begin{equation}
{\bm{A}_M}_{ij} =
    \begin{cases}
    -\inf & \text{if $\bm{A^{'}}_{ij} \geq \alpha $ } \\
    0, & \text{if $\bm{A^{'}}_{ij} < \alpha $}
    \end{cases}
    \label{eq:get_mask}
\end{equation}
The motion region attended to by the TrackAdapter is the part of the attention map $\bm{A^{'}}$ that exceeds $\alpha$. By setting the equivalent area in the attention mask $\bm{A_{M}}$ to $-inf$, the original temporal self-attention will no longer pay attention to this part and produce the separation effect. Then, the attention map of the original temporal self-attention can be rewritten as,
\begin{equation}
    \bm{A} = \mathrm{softmax}(\frac{\bm{Q}\bm{K}^{T}}{\sqrt{d}} + \bm{A}_M)
    \label{eq:origin}
\end{equation}
Finally, we get the temporal self-attention output of the current block: $\bm{O} = (\bm{A} + \bm{A}^{'})\bm{V} = \bm{A}^{''}\bm{V}$, where $ \bm{Q} = \bm{X}\bm{W}_{q}, \bm{K} = \bm{X}\bm{W}_{k}, \bm{V} = \bm{X}\bm{W}_{v}$ are the query, key, and value matrices of the temporal self-attention operation, $\bm{X}$ is input feature and $\bm{A''}$ is final attention map.
We complete the separation of the corresponding area and the unspecified area during the calculation of attention.

\subsection{Training and Inference of TrackAdapter}
\label{sec:train}

\subsubsection{Training}

The video diffusion model iteratively predicts noise $\epsilon $ in the noisy input, gradually transforming Gaussian noise into meaningful video frames.
The optimization of the model $\phi$ is achieved through noise prediction loss,
\begin{equation}
\mathcal{L}=\mathbb{E}_{t \sim \mathcal{U}(0,1), \epsilon \sim \mathcal{N}(\bm{0}, \bm{I})}
\left\|\epsilon_{\theta}\left( \bm{z_{t}} ; c\right)-\epsilon\right\|_{2}^{2}
\label{eq:base}
\end{equation}
\noindent where $t$ represents the timesteps, $\theta$ represents the U-Net's parameters, $c$ represents conditions and $\bm{z_t}$ is a noisily transformed version of ground truth video $\bm{z_{0}}$:
\begin{equation}
\bm{z_{t}} = \alpha_{t}\bm{z_{0}}+\sigma_{t}\epsilon
\end{equation}
Here, $\alpha_{t}$ and $\sigma_{t}$ denotes a predefined constant sequence. On the basis of Eq~\ref{eq:base}, we add point trajectories $\bm{P}$ and an image $\bm{I}$ as conditions, and the optimization objective can be written as,
\begin{equation}
\mathcal{L}_{\mathrm{mse}} = \left\|\epsilon_{\theta}\left( \bm{z}_{t} ; \mathcal{E}(\bm{P}) , \bm{I}, t \right)-\epsilon\right\|_{2}^{2}
\end{equation}
In order to adapt the original temporal self-attention to the new input mode quickly and to accelerate the model's convergence, we design an attention map based loss function. We gather attention maps from different temporal self-attention layers to get a set $\mathcal{C}$, which contains 16 different attention maps.
For $\bm{A}_q \in \mathcal{C}$, $\bm{A}_q(x,y)$ denotes the temporal self-attention map between frame $x$ and frame $y$ at block $q$. The purpose of the attention loss is to suppress the area corresponding to the mask in the original branch's attention map, \textit{i.e.} the motion area,
\begin{equation}
    \mathcal{L}_{\mathrm{attn}} = \sum_{\bm{A}_q \in \mathcal{C}}\sum_{i=1}^l (\bm{A}_q(i,i) * \phi(\bm{M}_i))^2
\end{equation}
where $\phi$ denotes the down-sampling operation, and $\bm{M_{i}}$ denotes the mask of all moving components in $i$-th frame. In total, our final loss function is then defined by the weighted average of the two terms,
\begin{equation}
\underset{\theta}{\operatorname{min}}  \; \nabla_{\theta} \mathcal{L}_{\mathrm{mse}} 
+ 
\lambda \mathcal{L}_{\mathrm{attn}} 
\end{equation}
\noindent where $\lambda$ is a hyperparameter.

\subsubsection{Inference.}
During inference, we set the intensity of the unspecified area to $\tau$, that is, we set the part of the Eq ~\ref{eq:get_mask} less than $\alpha$ to $\tau$.
Users can adjust $\tau$ to control the movement of the unspecified area in cases where it needs to move synchronously with the foreground movement or when sensory interference from the unspecified area needs to be mitigated.
This feature will greatly enhance the creation of fluid and highly synchronized motion videos.

\begin{figure*}[t]
  \centering
  \includegraphics[width=0.96\textwidth]{ 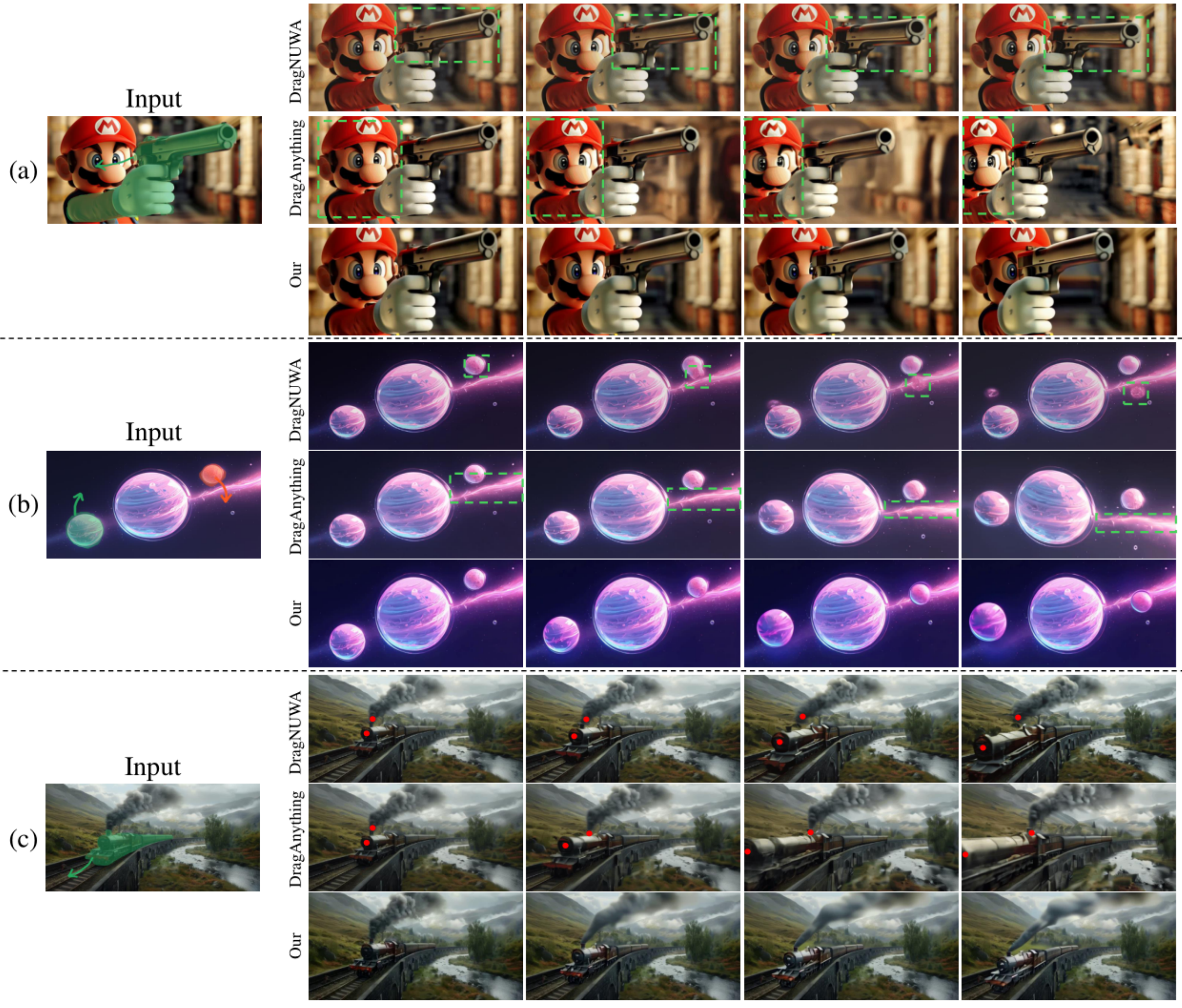}
  \caption{
    Qualitative comparisons between our method and baseline methods, DragAnything and DragNUWA. 
    We use colorful symbols to highlight the undesired parts of the results generated by the other two approaches.
    }
  \label{fig:compare}
\end{figure*}

\section{Experiment Settings}

\paragraph{Implementation Details.}

We employ SVD~\cite{blattmann2023stable} as our base model. All experiments were conducted using PyTorch with 8 NVIDIA A100-80G GPUs. AdamW~\cite{loshchilov2017decoupled} is configured as our optimizer, running for a total of 18,000 training steps with a learning rate of 3e-5 and a batch size of 8.
Following the method proposed in Animate Anyone~\cite{hu2023animate}, we have developed a lightweight encoder $\mathcal{E}$. This encoder employs a total of six convolution layers, two pooling layers, and a final fully-connected layer. Its primary function is to align the point trajectories $\bm{P}$ to the appropriate resolution. The query and key matrices of the TrackAdapter are initialized from the original temporal attention branch.

\begin{table*}[h]
\centering
\renewcommand{\arraystretch}{1.2}
\begin{tabular}{c@{\hspace{4pt}}|c@{\hspace{12pt}}c@{\hspace{4pt}}c@{\hspace{4pt}}|c@{\hspace{12pt}}c@{\hspace{6pt}}c@{\hspace{4pt}}|c@{\hspace{6pt}}c@{\hspace{4pt}}}
\toprule
 & \multicolumn{3}{c@{\hspace{10pt}}|}{VIPSeg} & \multicolumn{3}{c@{\hspace{10pt}}|}{Internal validation dataset} & \multicolumn{2}{c}{Performance Metrics} \\ 
Method & FVD $\downarrow$ & FID $\downarrow$ & ObjMC $\downarrow$ & FVD $\downarrow$ & FID $\downarrow$ & ObjMC $\downarrow$ & Parameters & Inference Time \\ 
\midrule
DragNUWA & 321.31 & 30.15 & 298.98 & 178.37 & 38.07 & 129.80 & 160.38M & 58.12s \\

DragAnything  & 294.91 & 28.16 & 236.02 & 169.73 & 32.85 & 133.89 & 685.06M & 152.98s \\

TrackGo  & \textbf{248.27} & \textbf{25.60} & \textbf{191.15} & \textbf{136.11} & \textbf{29.19} & \textbf{79.52} & \textbf{29.36M} & \textbf{33.94s} \\
\bottomrule
\end{tabular}
\caption{Quantitative comparisons of our approach with DragAnything and DragNUWA. All three methods are based on the same basic model called SVD. \textbf{Bolded values} indicate the best scores in each column.}
\label{tab:sampletable}
\end{table*}

\subsubsection{Dataset.}

For our experiments, we utilized an internal dataset characterized by superior video quality, comprising about 200K video clips. Following the experimental design, we further filtered the data to obtain a subset of about 110K videos as our final training dataset. During the training process, each video was resized to a resolution of $1024 \times 576$ and standardized to 25 frames per clip.

Our test set comprised the VIPSeg validation set along with an additional 300-video subset from our internal validation dataset.
Notably, all videos in the VIPSeg dataset are formatted in a 16:9 aspect ratio.
To maintain consistency, we adjusted the resolution of all videos in the validation sets to $1024 \times 576$. For evaluation purposes, we extracted the trajectories from first 14 frames of each video in the test set.

\paragraph{Evaluation Metrics and Baseline Methods.}

We measure video quality using FVD~\cite{unterthiner2018towards} and image quality using FID \cite{heusel2017gans}. We compare our methods with DragNUWA and DragAnything, which can also use trajectory information as conditional input. Following DragAnything, ObjMC is used to evaluate the motion control performance by computing the Euclidean distance between the predicted and ground truth trajectories.

\subsection{Quantitative Evaluation}
Quantitative comparisons of our method with baseline methods are shown in Table~\ref{tab:sampletable}. We test all models on VIPSeg validation set and internal validation set. From the results, TrackGo outperforms other approaches across all metrics, producing videos with higher visual quality and better fidelity to input motion control.
We also compared the model parameters and inference speed of the three approaches. Since all use the same base model, our comparison focused on the weights of the newly added modules.
To assess the model’s inference speed, we conducted 100 inference tests for each approach using identical input data on an NVIDIA A100 GPU. 
The results demonstrate that our approach not only delivers the best visual quality but also achieves the fastest inference speed, all while requiring the fewest additional parameters.

\subsection{Qualitative Evaluation}
\label{Qualitative Evaluation}

\subsubsection{Visualizations.}
We present a visualization comparison with DragAnything and DragNUWA, as shown in Fig.~\ref{fig:compare}.
We can have the following observations:
\textbf{First}, DragNUWA struggles with perceiving the control area, which may lead to incomplete or inaccurate optical flow. In case (b), the planet is not correctly perceived, and in case (a), the movement of the gun is also incorrect. In case (c), while the optical flow of the train is successfully predicted, the absence of flow in the smoke creates a jarring effect. Second, DragAnything struggles with partial or fine-grained object movement. In case (a), only the gun and Mario’s hand should move, but Mario’s entire position shifts unexpectedly, with a similar issue in case (b). DragAnything also fails to produce a harmonious background; in case (c), the smoke doesn’t follow the moving train. In contrast, TrackGo generates videos where target region movement aligns precisely with user input while maintaining background consistency and harmony. This significantly enhances the visual quality and coherence of the videos, showcasing TrackGo’s effectiveness. Additional examples are shown in Fig.~\ref{fig:tears}.

\subsubsection{Attention Mask Control of Background Movement.}
\begin{figure}[h]
  \centering
  \includegraphics[width=0.46\textwidth]{ 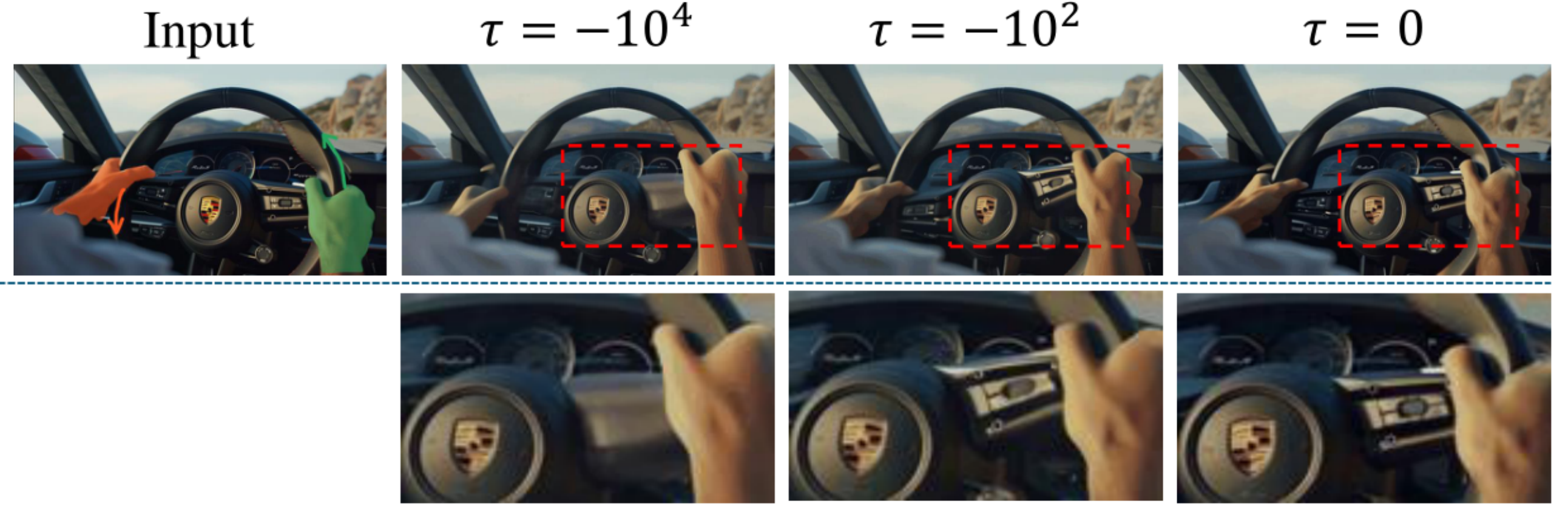}
  \caption{
    Comparison results of unspecified area suppression intensity $\tau$. The top row shows the results of the last frame generated for various $\tau$. The bottom row provides a magnified view, highlighting the differences more clearly with red boxes, to better observe the variations.
    }
  \label{fig:mask}
\end{figure}

Our model possesses the capability to adjust the intensity of motion in  unspecified areas through specific parameters, which is shown in the Fig.~\ref{fig:mask}. We define the motor inhibition intensity in unspecified areas with the parameter $\tau$.
As shown in Fig.~\ref{fig:mask}, when the hands move, it is necessary for the steering wheel to move correspondingly to enhance the realism of the output. When $\tau = -10^4$, the motion in unspecified areas is significantly suppressed, allowing only the hands to follow their trajectory while other parts remain static. This often results in distortion.
When $\tau = -10^2$, the unspecified area is less suppressed, allowing hands and the steering wheel to move simultaneously. However, this setting can still produce disharmonious outcomes, such as the car logo on the steering wheel remaining static.  
When $\tau = 0$, the unspecified areas can move freely, which typically results in a more cohesive and harmonious video. Nevertheless, not all movements in the unspecified areas are desirable, and excessive movement can damage the video quality. Therefore, it is crucial to carefully manage the suppression level to balance realism with artistic control.

\begin{table}[h]
    \centering
    \renewcommand{\arraystretch}{1.1}
    \setlength{\tabcolsep}{4pt}
    \begin{tabular}{@{}lccc@{}}
        \toprule
        \textbf{Train Step} & \textbf{14k} & \textbf{16k} & \textbf{18k} \\
        \midrule
        w/o Attn Mask and Loss  & 219.15 & 208.31 & 218.50 \\
        w/o Attn Loss  & 216.54 & 191.54 & 165.12 \\
        Full Method & \textbf{204.02} & \textbf{184.03} & \textbf{136.11} \\
        \bottomrule
    \end{tabular}
    \caption{FVD tested on the validation set of different variants after 14k, 16k, and 18k training steps.}
    \label{tab:ablation_results}
\end{table}

\subsection{Ablation Study}

To validate the effectiveness of the attention mask and attention loss, we report the FVD metrics on the internal validation set at various training steps, as shown in Table~\ref{tab:ablation_results}. Under the same number of training steps, the model without attention loss shows a slightly higher FVD compared to the model with attention loss. When attention loss is not utilized, the FVD is higher compared to when attention loss is applied. This discrepancy becomes particularly pronounced at the 18K training mark. This demonstrates that using attention loss can accelerate model training and aid convergence. Without both the attention mask and attention loss, the FVD stabilizes around 16K steps but remains significantly higher than the FVD under the full setting.

\section{Conclusion}

In this paper, we introduce point trajectories to capture complex temporal information in videos. We propose the TrackAdapter to process these point trajectories, focusing on the motion of specified targets, and employ an attention mask to mitigate the influence of original temporal self-attention on specified regions. During inference, the attention mask can regulate the movement of unspecified areas, resulting in video output that aligns more closely with user input. Extensive experiments demonstrate that our TrackGo achieves state-of-the-art FVD, FID, and ObjMC scores.

\section{Acknowledgments}

This work was supported by the National Science and Technology Major Project (No. 2022ZD0117800), Young Elite Scientists Sponsorship Program by CAST, and the Fundamental Research Funds for the Central Universities.

\bibliography{aaai25}

\clearpage

\twocolumn[
    \begin{center}
        {\LARGE \bfseries Supplementary Material}
    \end{center}
]


\section*{Overview}
\label{sec:overview}

This supplementary material is organized into several sections, each providing additional details and analyses related to our work on TrackGo. The sections are structured as follows:

\begin{itemize}
\item In Section A, we present the implementation details of TrackGo, including the default parameters used in the pipeline.
\item In Section B, we provide the experimental settings and details of our internal dataset.
\item In Section C, we conduct a user study to demonstrate the popularity of our method compared to existing approaches.
\item In Section D, we compare our method with the Stable Video Diffusion Model (SVD)~\cite{blattmann2023stable}, which serves as the baseline for our method, thereby demonstrating the effectiveness of our proposed TrackAdapter.
\item In Section E, we present additional qualitative results of TrackGo.
\item In Section F, we discuss the limitations of the current version of TrackGo and propose potential improvements.
\item In Section G, we examine the societal impact of TrackGo.
\end{itemize}

\section{A. Implementation Details of TrackGo}

\noindent\textbf{How to Determine $K$.}\quad When generating point trajectories from the training dataset, we set the maximum number of significant components, $s$, to 6. The number of points $K$, which ranges from 16 to 64, is determined by the area of the component's mask, as shown in Eq.~\ref{eq:point_num}.
\begin{equation}
K = 
\begin{cases} 
16 & \text{if } p < 0.1 \\
16 + \frac{(64 - 16)}{(0.4 - 0.1)} \cdot (p - 0.1) & \text{if } 0.1 \leq p \leq 0.4 \\
64 & \text{if } p > 0.4 
\end{cases}
\label{eq:point_num}
\end{equation}
where $p$ represents the proportion of the component's area to the total image area.

\noindent\textbf{How to Set the Colors of the Control Points.}\quad For each control point, we create a visual representation by drawing a square with a 4-pixel side length centered at its coordinates. Colors for these squares are assigned according to a predefined list, correlating with the component each belongs to, and following a specific drawing order.

\noindent\textbf{How to Obtain the Attention Mask.}\quad We insert the TrackAdapter into the temporal self-attention of each layer of the 3D U-Net. We obtained the attention map $\bm{A'}$ from TrackAdapter and derived a binarized mask through the threshold $\alpha$, as shown in Eq.~\ref{eq:get_mask}. The $\alpha$ is set to $0.2$.
\begin{equation}
{\bm{A}_M} =
    \begin{cases}
    -\inf & \text{if $\bm{A^{'}} \geq \alpha $ } \\
    0, & \text{if $\bm{A^{'}} < \alpha $}
    \end{cases}
    \label{eq:get_mask}
\end{equation}

To prevent overflow during computation with the \textit{torch.fp16} data type, we set the value of attention map $\bm{A_{M}}$ to $-10^4$ instead of $-inf$.

\section{B. Experimental Settings and Dataset}

In this section, we provide additional details about TrackGo's experimental environment and the internal dataset.

All experiments are performed on a server with 8 NVIDIA A100 GPUs and Intel(R) Xeon(R) Platinum 8336C CPUs. We used the 80GB version of the NVIDIA A100 GPU. During model inference, it takes about 30 seconds to generate a video.

The internal dataset we used to train our model contains approximately 200K video clips, which vary in length from 20 to 600 frames, with a resolution of $1280 \times 720$. Most of the video content is real-world. These videos are of high quality and include a variety of motion types, such as vehicle movements, human actions, and various activities. 
We removed videos from the dataset that were of relatively low quality, had minimal movements, contained discontinuous segments (with transitions or frame jumps), or exhibited significant camera motion. Ultimately, we obtained about 110K video clips as our training dataset.

\begin{figure}[hbt]
  \centering
  \includegraphics[width=0.45\textwidth]{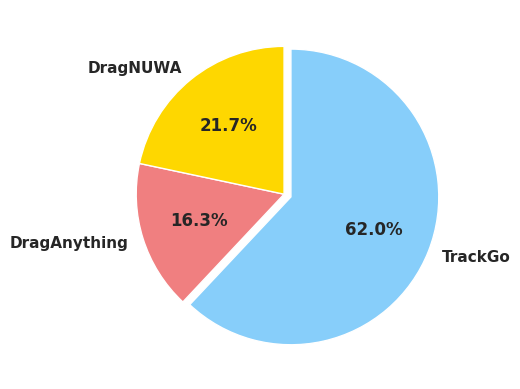}
  \caption{
    The results of the user study.
    }
  \label{fig:user_study}
\end{figure}

\begin{figure*}[bt]
  \centering
  \includegraphics[width=0.92\textwidth]{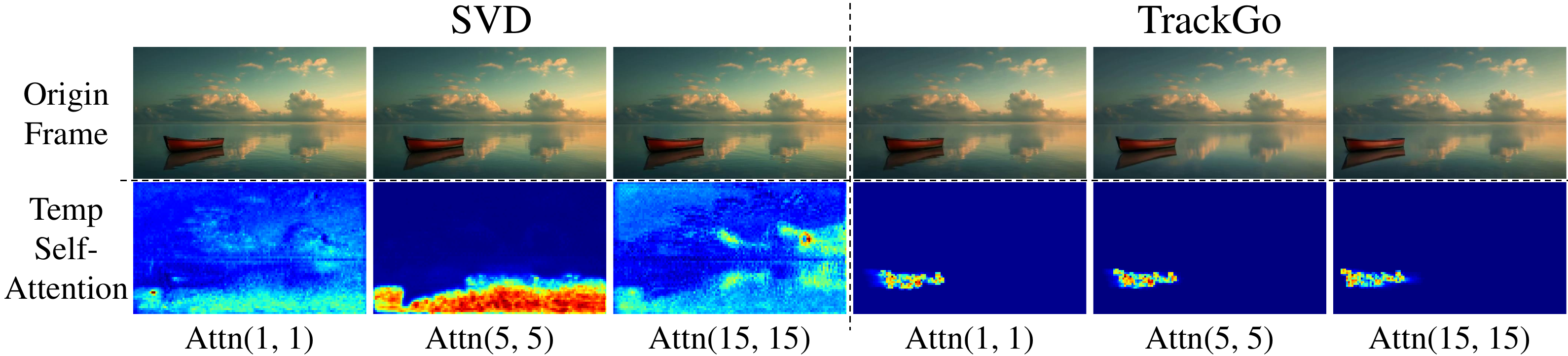}
  \caption{
  Visualizations of the  temporal attention map of SVD and TrackGo.
  Both approaches use the same image as the input, and the parameters used for inference are the same. With our proposed TrackAdpater, the temporal self-attention layers of TrackGo can better focus on the intended moving object.
  }
  \label{fig:compare_svd}
  \vspace{-0.2cm}
\end{figure*}

\section{C. User Study}

We conducted a user study to assess the quality of the synthesized videos. We randomly sampled 60 cases, with the results of three different approaches for user study. Each questionnaire contains 30 cases which are randomly sampled from these 60 cases. 
We asked the users to choose the best based on overall quality in terms of two aspects: the consistency between the generated video and the given conditions, and the quality of the generated video (i.e., whether the subject is distorted, whether the unselected background is shaky, \textit{etc.}).

We invited 30 people to fill out the questionnaire, with a gender ratio of approximately 3:1 (Male: Female). Most of the participants are university students from various fields of science and engineering, ranging in age from 18 to 27.
The results show that our approach achieved 62\% of the votes, higher than DragAnything's 16.33\% and DragNUWA's 21.67\%, as shown in Fig.~\ref{fig:user_study}.

\section{D. Comparison with SVD}

In this section, we compared TrackGo to its baseline SVD. We show the differences in the results generated by TrackGo and SVD to demonstrate the effectiveness of TrackAdapter.

As shown in Fig.~\ref{fig:compare_svd},  without TrackAdapter, the SVD's results focus more on the dynamic effect of the water surface, leaving the ship's motion inconspicuous. 
In our TrackGo, users used the masks and arrows to explicitly indicate the moving objects and trajectories. With the help of TrackAdapter, the temporal self-attention layers can focus on the intended area, \textit{i.e.}, the ship.

\section{E. More Qualitative Results}

\begin{figure*}[htb]
  \centering
  \includegraphics[width=0.95\textwidth]{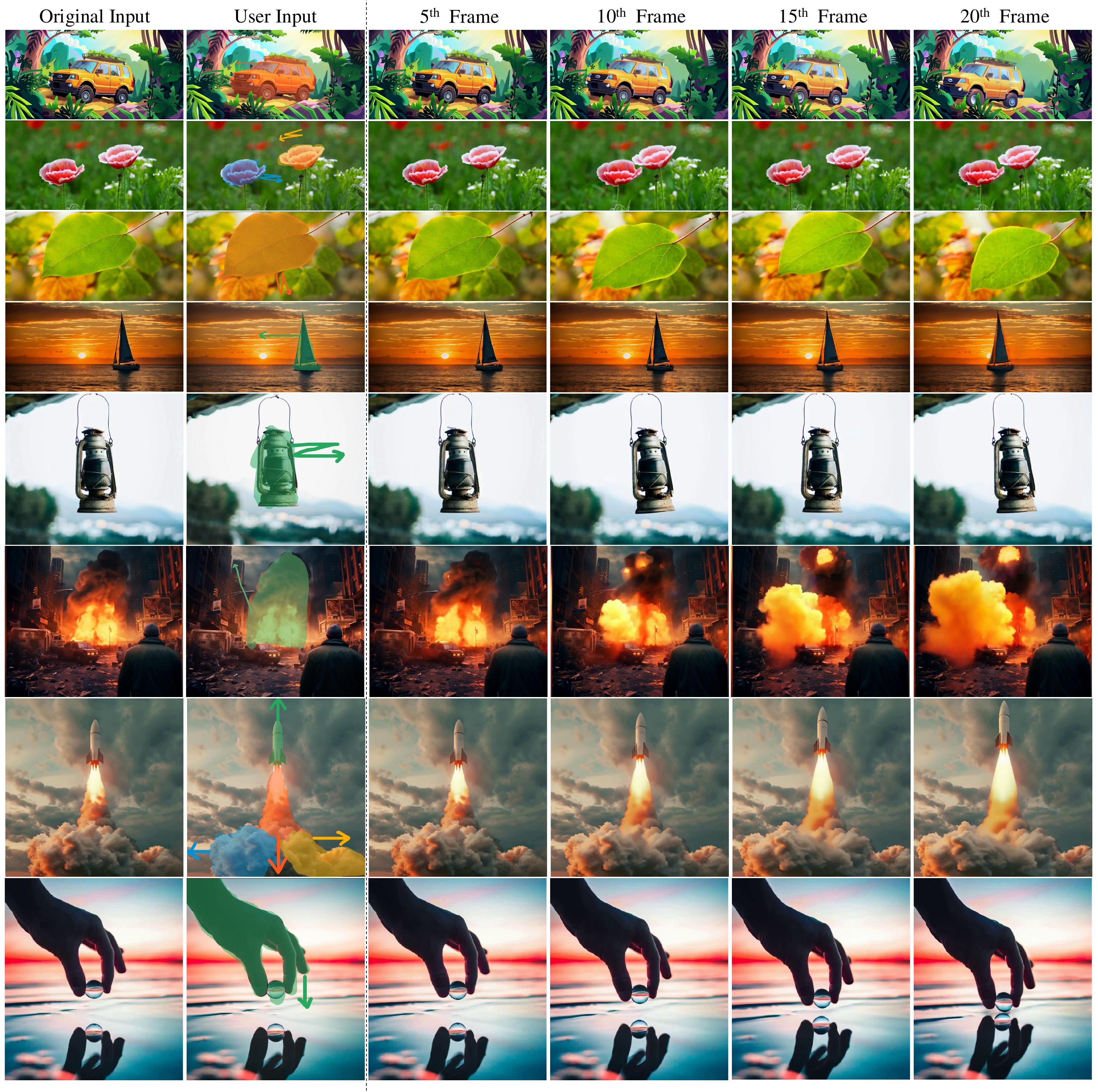}
  \caption{
    More qualitative results generated by TrackGo.
    }
  \label{fig:more}
\end{figure*}

In Fig.~\ref{fig:more}, we provide more qualitative results from TrackGo. 

\paragraph{Examples of Diversity.}

\begin{figure*}[htb]
  \centering
  \includegraphics[width=1\textwidth]{ 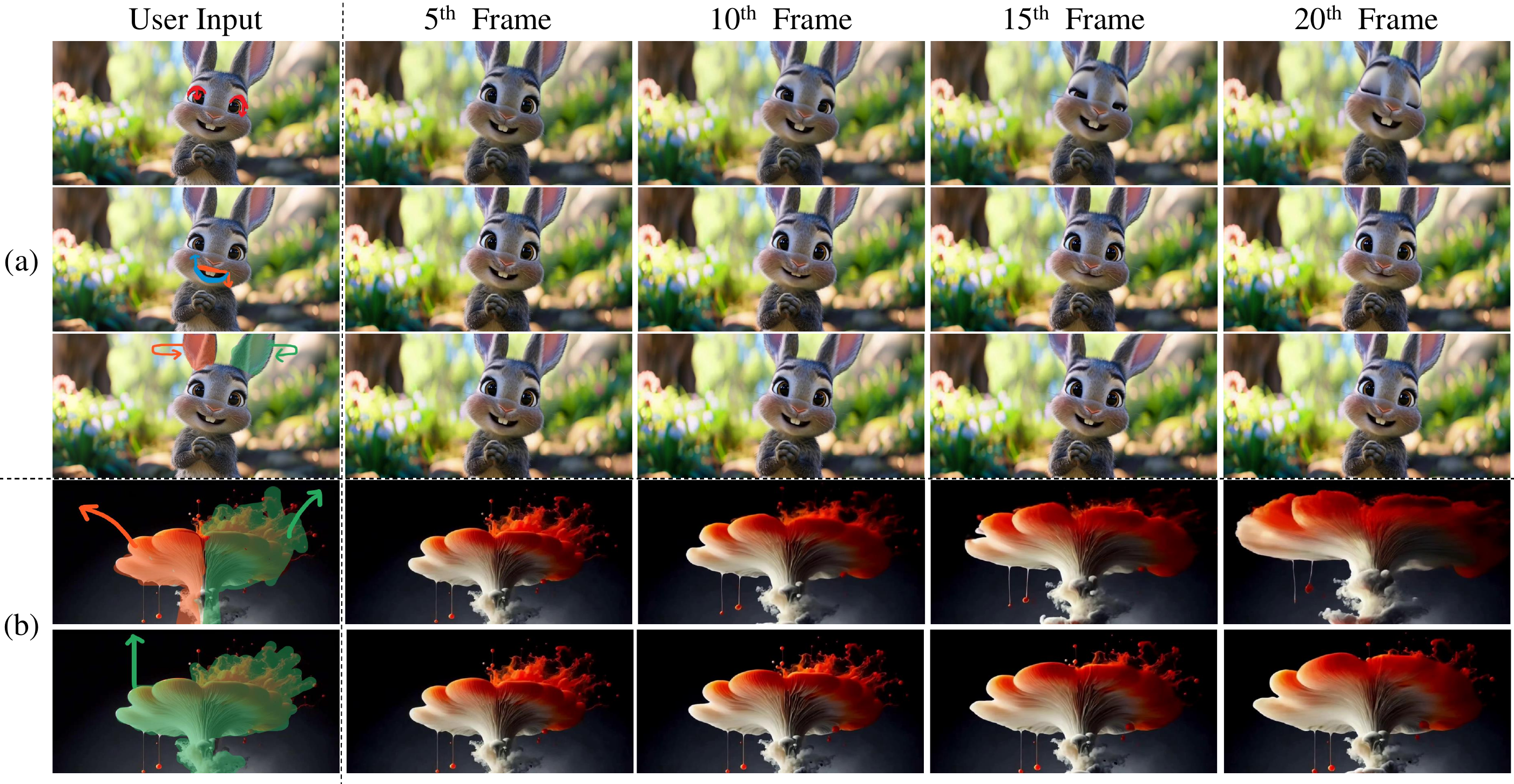}
  \caption{
    Various results of TrackGo. \textbf{(a)} For the same input image, TrackGo can generate different video clips with different user input. 
    \textbf{(b)} TrackGo is robust to user inputs. TrackGo can realize similar moving effects with different user inputs.
    }
  \label{fig:rabbit}
\end{figure*}

As shown in Fig~\ref{fig:rabbit}, TrackGo can achieve diverse results based on the same image using different control conditions. For instance, users can choose to keep the rabbit's eyes or mouth closed or let the rabbit's ears swing back and forth. Under these varying control conditions, TrackGo demonstrates good diversity in its outputs.
Besides, TrackGo can generate similar outputs with different user input conditions, indicating our model is robust to user inputs.

\begin{figure*}[htb]
  \centering
  \includegraphics[width=1\textwidth]{ 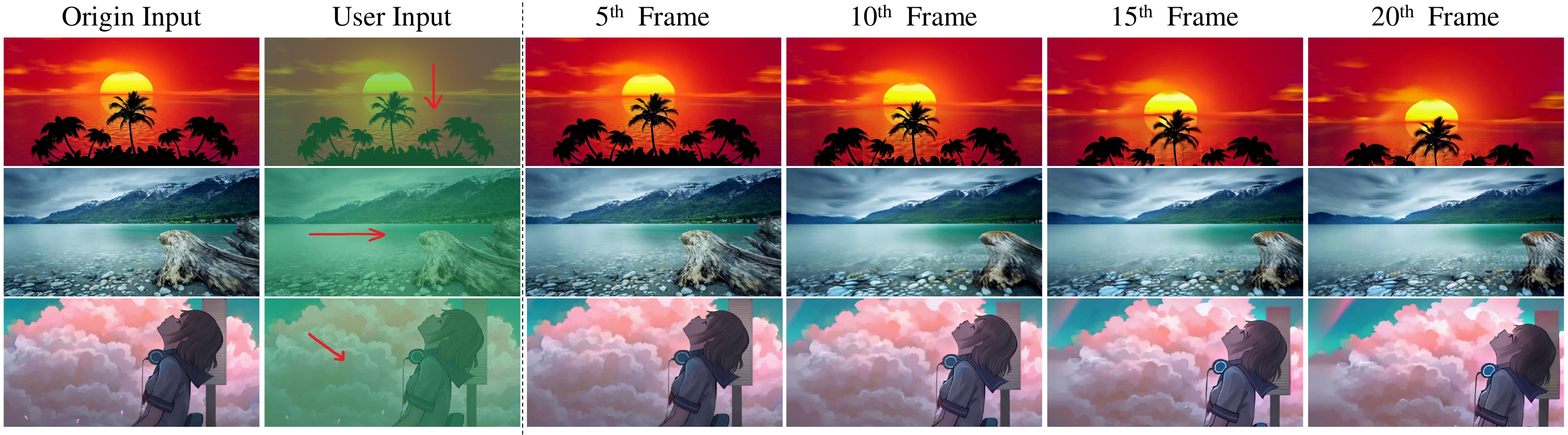}
  \caption{
    Results generated by TrackGo with  camera control. Note that to realize camera motion, the entire image is selected as the motion area.
    }
  \label{fig:camera}
\end{figure*}

\paragraph{Camera Motion.}

Like DragAnything\cite{wu2024draganything}, TrackGo can also achieve the effect of camera motion, as shown in Fig~\ref{fig:camera}. By simply selecting the entire image region as the motion area and providing a trajectory, an effect where the camera moves in the specified direction of the trajectory can be achieved.

\begin{figure*}[htb]
  \centering
  \includegraphics[width=1\textwidth]{ 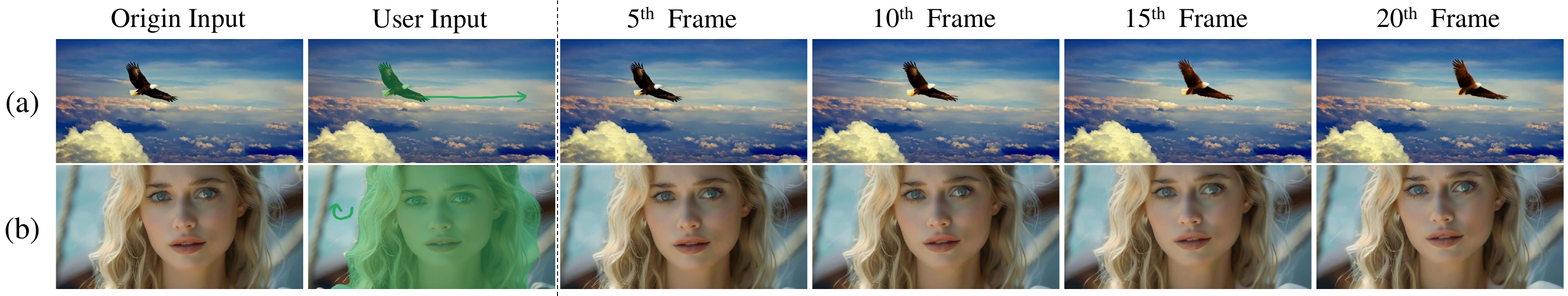}
  \caption{
    Failure cases of TrackGo.
    }
  \label{fig:failure_case}
\end{figure*}

\section{F. Limitations}

Our approach has two main limitations. Specifically, our model struggles to deal with large movements. Some details may be distorted in generated video clips. As shown in Fig.~\ref{fig:failure_case}, the bird's head is distorted in the last few frames.
This limitation can be mitigated by further enhancing the capabilities of the foundational model. 
A possible improvement is implementing our model based on the DiT\cite{peebles2023scalable} architecture, such as Sora\cite{Sora}, to gain improved motion prior knowledge.


Another limitation is that our model cannot handle 3D rotation information, such as rotating a face. The model often interprets rotation as translation, generating incorrect results, as shown in case (b) of Fig.~\ref{fig:failure_case}. 
To address this issue, we could incorporate 3D rotation information to help the model better understand 3D rotational motion patterns, allowing it to generate videos that depict more complex motion.

\section{G. Societal Impact}

TrackGo improves the controllability of video generation. However, as with any generative model, there is a critical issue regarding the potential negative impact of our model being used to create fake videos on society. We must make concerted efforts to ensure that the model is used responsibly to avoid causing negative social impacts.

\end{document}